\documentclass[a4paper,11pt,preprint]{elsarticle}
 
\pdfoutput=1

\usepackage{lineno,hyperref}
\usepackage{amsmath}
\usepackage{mathtools}
\usepackage{amssymb} 
\usepackage{amsthm}
\usepackage{bbm}
\usepackage{graphicx,epsf}
\usepackage{enumerate}
\usepackage{subcaption}
\usepackage{caption}
\usepackage{array}
\usepackage{natbib}
\usepackage{bm}
\usepackage{color}
\modulolinenumbers[5]
\graphicspath{{figures/}} % comment out for arXiv

\DeclareMathOperator*{\argmin}{arg\,min}

\DeclareMathOperator{\woe}{\mbox{WOE}}
\DeclareMathOperator{\swoe}{\mbox{SWOE}}
\DeclareMathOperator{\cwoe}{\mbox{CWOE}}
\newcommand{\bx}{\bm x}
\newcommand{\bX}{\bm X}
\newcommand{\bb}{\bm \beta}

\newcommand{\checkedcb}{\makebox[0pt][l]{$\square$}\raisebox{.15ex}{\hspace{0.1em}$\checkmark$}}
\newcommand{\emptycb}{\makebox[0pt][l]{$\square$}\raisebox{.15ex}}

\journal{arXiv:2101.01494}

\begin{document}

\begin{frontmatter}

\title{Weight-of-evidence through shrinkage and spline binning for interpretable nonlinear classification}
% \tnotetext[mytitlenote]{Fully documented templates are available in the elsarticle package on \href{http://www.ctan.org/tex-archive/macros/latex/contrib/elsarticle}{CTAN}.}
%\shorttile{Weight-of-evidence 2.0}
%% Group authors per affiliation:
% \author{\fnref{myfootnote}}
% \address{Radarweg 29, Amsterdam}
% \fntext[myfootnote]{Since 1880.}

\author{Jakob Raymaekers}
\address{Department of Mathematics, KU Leuven, Leuven, Belgium}
\address{Department of Mathematics, University of Antwerp, Antwerp, Belgium}

\author{Wouter Verbeke}
\address{Faculty of Economics and Business, KU Leuven, Leuven, Belgium}

\author{Tim Verdonck\fnref{mycorrespondingauthor}}
\fntext[mycorrespondingauthor]{Corresponding author. E-mail: tim.verdonck@uantwerpen.be}
\address{Department of Mathematics, University of Antwerp, Antwerp, Belgium}
\address{Department of Mathematics, KU Leuven, Leuven, Belgium}

% %% or include affiliations in footnotes:
% \author[mymainaddress,mysecondaryaddress]{Elsevier Inc}
% \ead[url]{www.elsevier.com}

% \author[mysecondaryaddress]{Global Customer Service\corref{mycorrespondingauthor}}
% \cortext[mycorrespondingauthor]{Corresponding author}
% \ead{support@elsevier.com}

% \address[mymainaddress]{1600 John F Kennedy Boulevard, Philadelphia}
% \address[mysecondaryaddress]{360 Park Avenue South, New York}

\begin{abstract}
In many practical applications, such as fraud detection, credit risk modeling or medical decision making, classification models for assigning instances to a predefined set of classes are required to be both precise and interpretable. Linear modeling methods such as logistic regression are often adopted since they offer an acceptable balance between precision and interpretability. Linear methods, however, are not well equipped to handle categorical predictors with high cardinality or to exploit nonlinear relations in the data. As a solution, data preprocessing methods such as weight of evidence are typically used for transforming the predictors. The binning procedure that underlies the weight-of-evidence approach, however, has been little researched and typically relies on ad hoc or expert-driven procedures. The objective in this paper, therefore, is to propose a formalized, data-driven and powerful method. To this end, we explore the discretization of continuous variables through the binning of spline functions, which allows for capturing nonlinear effects in predictor variables and yields highly interpretable predictors that take only a small number of discrete values. Moreover, we extend the weight-of-evidence approach and propose to estimate the proportions using shrinkage estimators. Together, this method offers an improved ability to exploit both nonlinear and categorical predictors to achieve increased classification precision while maintaining the interpretability of the resulting model and decreasing the risk of overfitting. We present the results of a series of experiments in fraud detection and credit risk settings, which illustrate the effectiveness of the presented approach.
\end{abstract}

\end{frontmatter}

%\linenumbers
\clearpage
%\tableofcontents
	
\section{Introduction}\label{sec:introduction}

Classification is a well-studied machine learning task that concerns the assignment of instances to a set of outcomes. Classification models support the optimization of managerial decision making across a variety of operational business processes. For instance, fraud detection models classify instances, such as transactions or claims, as fraudulent or nonfraudulent \cite{VANHOEYVELD2020105895}. This allows for the efficient and effective allocation of limited inspection capacity by selecting the most suspicious cases for investigation by a human fraud analyst \cite{baesens2015fraud}. Credit risk models, on the other hand, assess the risk connected with providing credit to customers, and this risk can be used to construct optimal portfolios of loans or other lines of credit \cite{baesens2016credit, bluhm2016introduction}.

A wide variety of classification models have been proposed in the literature. These proposals range from very complex models including neural networks, support vector machines and ensemble methods to more elementary models such as logistic regression and decision trees \cite{hastie2009elements}. Some of the more complex models have been shown to outperform the simpler classification techniques in various real-life classification tasks \cite{CHANG2018914,shi2012comparison, lessmann2015benchmarking, GUNNARSSON2021,  OSKARSDOTTIR201926}. In industry, however, simple logistic regression currently remains among the most frequently used approaches for developing classification models across various fields of application \cite{SOHN2016150, baesens2003benchmarking, lessmann2015benchmarking, GUNNARSSON2021, dastile2020statistical}.
Its popularity may be explained by the presence of industry regulations, e.g., the Basel regulatory framework for the banking industry, which requires the resulting model to be both interpretable \cite{martens2011performance} and accurate. Logistic regression is widely perceived as offering the best balance between both objectives. Other possible explanations are the broad expertise and experience in using logistic regression that exists in industry, but follow-the-herd behavior and some degree of inertia and resistance to change may explain its enduring popularity.
Moreover, the superior performance of the more complex models can strongly depend on the task at hand. On tabular datasets, which are commonly encountered in the context of credit scoring and healthcare analytics, they have been shown to provide only marginal performance gains \cite{lessmann2015benchmarking,rajkomar2018scalable}.

Aside from the development of classification models and techniques for learning them, a different approach to improving the final model is to focus on pre- and post-processing. In contrast to studies on learning models and post-processing techniques \cite{verbeke2017rulem, HERASYMOVYCH2019105697}, relatively few studies focus on preprocessing data. The goal of preprocessing is to 
 optimally prepare the data (e.g., through transformation) to maximize the predictive power and out-of-sample performance, or, importantly, to improve the interpretability of the resulting model. Specifically, we identify a lack of approaches that allow us to optimally transform nonlinear patterns and categorical variables with high cardinality for incorporation in linear models to achieve an interpretable yet powerful classifier \cite{moeyersoms2015including}. Currently, the weight-of-evidence (WOE) approach appears to be frequently used to this end, as it offers a good balance between interpretability and predictive power, and it is complementary with and similar to logistic regression \cite{smith2002weight,anderson2007credit}. For categorical variables with many categories, however, WOE may lead to overfitting. Moreover, WOE does not have an integrated binning approach for optimally merging categories or discretizing continuous predictors.

%In contrast to studies on learning classification models and post-processing techniques \cite{verbeke2017rulem, HERASYMOVYCH2019105697}, relatively few studies focus on preprocessing data, i.e., transforming the data prior to learning a classification model, and on developing approaches for optimally preparing the data to maximize their predictive power and out-of-sample performance, or, importantly, to improve the interpretability of the resulting model. Specifically, we identify a lack of approaches that allow us to optimally transform nonlinear patterns and categorical variables with high cardinality for incorporation within linear models to achieve an interpretable yet powerful classifier \cite{moeyersoms2015including}. Currently, the weight-of-evidence (WOE) approach appears to be frequently used to this end, as it offers a good balance between interpretability and predictive power, and it is complementary with and similar to logistic regression \cite{smith2002weight,anderson2007credit}. In cases with high cardinality, however, WOE may lead to overfitting. Moreover, WOE does not have an integrated binning approach for optimally merging categories or discretizing continuous predictors.

In this article, we present an integrated WOE-based approach for optimally transforming predictor variables, which mainly improves upon the existing WOE approach in cases with nonlinear predictor variables (continuous or ordinal) and categorical predictor variables with high cardinality. The goal of this preprocessing method is to maximize both the predictive power and interpretability of logistic regression models (and more generally, generalized linear models). The proposal is based on generalized additive models in combination with exact univariate $k$-means clustering and shrinkage estimation. The presented approach is experimentally evaluated; an illustration of the use of the proposed approach and an indication of its merits are provided. An open source implementation of the method is provided in the digital annex to this paper to enable peer researchers to reproduce and verify the presented results and allow practitioners to adopt the method for practical use.
This paper is structured as follows. In the following section, we present the standard methodology that uses logistic regression and weight-of-evidence, and we expand upon this approach in Section 3. In Section 4, we present experimental results obtained from a fraud detection case and a credit risk case, and in Section 5, we conclude the paper and present directions for future research.

%In this article, we present an integrated WOE-based approach for optimally transforming predictor variables, which mainly improves upon the existing WOE approach in cases with nonlinear predictor variables (continuous or ordinal) and categorical predictor variables with high cardinality, with the aim of maximizing both the predictive power and interpretability of (although not limited to these traits) logistic regression models. The presented approach is experimentally evaluated; an illustration of the use of the proposed approach and an indication of its merits are provided. An open source implementation of the method is provided in the digital annex to this paper to enable peer researchers to reproduce and verify the presented results and allow practitioners to adopt the method for practical use.
%This paper is structured as follows. In the following section, we present the standard methodology that uses logistic regression and weight-of-evidence, and we expand upon this approach in Section 3. In Section 4, we present experimental results obtained from a fraud detection case and a credit risk case, and in Section 5, we conclude the paper and present directions for future research.

\section{Background methodology}
Consider a model with a binary response $Y$ and $p$ continuous predictors $\bX = (X_1,\ldots,X_p)$. The goal is to model the conditional mean $p_{\bx} = E(Y|\bX =\bx) = P(Y = 1 | \bX =\bx)$. The classical logistic regression model, which is part of the family of generalized linear models (GLMs) \cite{nelder1972generalized}, assumes a linear relationship between the predictor variables and the log-odds of the event $Y=1$. More specifically, we have
\begin{equation*}
 \log\left(\frac{p_{\bx}}{1-p_{\bx}}\right) = \beta_0 + \sum_{i=1}^{p}{\beta_i x_i} = \beta_0 + \bb \bx
\end{equation*}
where $\beta_0$ denotes an intercept and $\bb = (\beta_1, \ldots, \beta_p)$ denotes a vector of model parameters. This model can be reformulated in terms of probabilities as
\begin{equation*}
P(Y = 1 | \bX =\bx) = \frac{1}{1+e^{-(\beta_0 + \bb \bx)}}.
\end{equation*}

The classical logistic regression model serves as a very popular benchmark for many binary classification tasks due to its ease of computation, high interpretability and solid performance. However, it also has several shortcomings, two of which we want to focus our attention on:
\begin{enumerate}
\item categorical variables with many categories
\item continuous variables with nonlinear effects on the log-odds
\end{enumerate}

Categorical variables are often one-hot encoded (also known as ``dummy encoding''), after which they can be included in the model as numerical variables. This has the drawback that a categorical variable with $N$ categories leads to $N - 1$ variables. If $N$ is large, this leads to considerable variability in the estimation process and usually many insignificant predictors. One way to avoid this problem is by converting the categorical variable into a continuous variable by using a weight-of-evidence transformation.
The 
%Editor: Abbreviations and acronyms typically need to be defined only once within the main text. Please consider adhering to this convention.
weight-of-evidence (WOE) 
transformation of a categorical predictor is commonly defined as follows. Suppose that we have a category $j$ with $N_j$ elements. Denote by $P_j$ the number of true cases in our category and by $F_j$ the number of false cases in our category. Additionally, let $P$ be the total number of true cases in the data and $F$ be the total number of false cases in the data.
The WOE value of category $j$ is then given by:
\begin{equation}\label{eq:woe1}
\log\left(\frac{P_j / P}{F_j / F}\right).
\end{equation}
The WOE transformation usually provides an elegant solution, but since it is based on the estimation of a proportion, its variance can be high when there are categories with few observations, which is common for categorical variables with high cardinality.

Continuous variables are modeled by logistic regression as having linear effects on the log-odds of the response. While this is often reasonable, there can be variables that do not satisfy this assumption. This happens unexpectedly but sometimes by design, as illustrated in the following example. Suppose that we want to predict whether a transaction is fraudulent based on a single predictor $X_t$ that characterizes the time at which the transaction was made (i.e., taking values within $[0, 24)$). Now suppose that we make the reasonable assumption that the influence of the time on the probability of a transaction being fraudulent is roughly continuous, and we interpret $X_t$ being close to 24 as $X_t$ being close to 0. Then, we would have $P(Y = 1 | X_t = 0) = \displaystyle \lim_{T \to 24^-}{P(Y = 1 | X_t = T)}$. In terms of log-odds, this would imply that $\beta_0 =  \displaystyle  \lim_{T \to 24^-}{\beta_0 + \beta_1 T} = \beta_0 + 24\beta_1$, which clearly can only be satisfied when $\beta_1 = 0$. In other words, under the assumptions above, the only relationship that can be fit is a constant relationship, which is not of much interest. This example illustrates that some variables display nonlinear relationships with the response by design.\par
One way to incorporate the nonlinear effects of continuous predictors on the log-odds is to use the generalized additive model (GAM, \cite{hastie1987generalized, WoodGam}) for logistic regression:
\begin{equation}\label{eq:gam}
 \log\left(\frac{p_{\bx}}{1-p_{\bx}}\right) = \beta_0 +\sum_{i=1}^{p}{f_i(x_i)}
\end{equation}
where $f_1,\ldots,f_p$ are arbitrary smooth functions of the predictor variables $x_1,\ldots,x_j$.
The model in Eq. ~\ref{eq:gam} is very flexible, but this flexibility comes at a price. As the functions $f_i$ can be arbitrary smooth functions of the predictors, they can display rather unusual patterns. These factors make the model harder to interpret and hence less used in practical situations such as fraud detection, where the predictions resulting from the model may have to be explained. To improve the interpretability of the model, \cite{Henckaerts2018} proposed a data-driven way of binning the fitted functions $f_i$ into a limited number of categories. Afterwards, a classical logistic regression model can be fit to the binned variable. This strategy allows for capturing nonlinear effects while greatly improving the interpretability of the model.\\

Throughout the remainder of the article, we make the assumption that the conditional expectation of $Y$ can indeed be adequately modeled through a GAM in the predictor variables. This assumption entails that the nonlinear effects are sufficiently smooth functions of the predictors. Furthermore, we assume that the number of variables $p$ is considerably smaller than the number of observations $n$ which guarantees stability in fitting the model. In case this last assumption would not be met, one could resort to regularized GAMs and apply the propsed methodology in that setting.

\section{Methodology}\label{sec:meth}
In the following, we describe our proposal to address the issues described in the previous section. The underlying goal is to develop a powerful predictive model while maintaining interpretability by allowing the incorporation of nonlinear effects within a GLM and by improving upon the traditional WOE-based binning process.

\subsection{(Local) shrinkage of WOE}

Our starting point for the treatment of categorical variables is the WOE transformation that transforms a categorical variable into continuous values. To introduce our shrinkage estimator for the WOE values, we first rewrite the definition of Eq. ~\ref{eq:woe1} in a different but equivalent form. More specifically, for a given categorical variable, we assign the empirical log-odds to each bin, i.e., each element in a given category $j$ is assigned the value
\begin{equation}\label{eq:woe2}
\widehat{\mbox{WOE}_j} = \log\left(\frac{\hat{p}_j}{1-\hat{p}_j}\right)
\end{equation}where $\hat{p}_j$ denotes the proportion of successes (e.g., fraudulent transactions) in category $j$. The equivalence with the earlier definition in Eq.~\ref{eq:woe1} can be seen as follows. With the notation introduced before, we have that $\mbox{WOE}_j = \log\left(\frac{p_j}{1-p_j}\right) =  \log\left(\frac{P_j / N_j}{F_j / N_j}\right) = \log\left(\frac{P_j}{F_j}\right)= \log\left(\frac{P_j / P}{F_j / F}\right) + \log\left(\frac{P}{F}\right)$. Therefore, both values differ by only a constant, which typically does not play a role in most statistical or machine learning models. As an example, the constant disappears in the intercept of a GLM. It is worth noting that sometimes categories with $\hat{p} = 0$ or $\hat{p} = 1$ can occur, and these lead to undefined WOE values. In those cases, we can slightly adjust the WOE by introducing a small offset $c$ with $0 < c < 1$ and replace $\hat{p} = 0$ with $\hat{p} =\frac{c}{n_j}$ and $\hat{p} = 1$ with $\hat{p} =1 - \frac{c}{n_j}$. Note that this offset disappears as the number of observations in the category becomes large (i.e., when $n_j \to \infty$). We use $c = 0.01$ by default. In practice, categories are often merged to avoid this boundary case, but this merging introduces a certain level of arbitrariness. In particular, it raises the question as to whether all possible combinations of categories should be considered as possible merging candidates. Additionally, this technique does not use the performance or quality of the final model for evaluating which merges are most interesting. We thus prefer working with a small offset, after which we can deal with the WOE values in a rigorous way.

For a category with a small number of observations $n_j$, the estimation of $p_j$ (and the corresponding $\mbox{WOE}_j$) has a high variance, often yielding unreliable estimates. This is more likely to occur in categorical variables with many levels. To address this issue, we consider the shrinkage estimation of the proportion of successes in each category $j$. The shrinkage estimator of a proportion is given by \cite{longford1999multivariate}:
$$\tilde{p}_j = (1-b_j)\hat{p}_j + b_j \hat{p}$$
where $\hat{p}$ denotes the proportion of successes calculated over all possible values of $j$ (i.e., over all categories). We thus effectively shrink the proportion of successes towards the sample mean. The shrinkage coefficient $b_j$ determines the amount of shrinkage: $b_j = 0$ corresponds to no shrinkage, whereas $b_j= 1$ corresponds to taking the population proportion. The value of $b_j$ is chosen to minimize the expected mean squared error over all estimated proportions, which is given by $\mbox{EMSE} = E_s[E_j[(\tilde{p}_j-p_j)^2|p_j]$. The minimum is given by (provided $n_j/n < 0.5$):
$$b_j^* = \frac{v_j(1 - n_j/n)}{v_j(1 - 2n_j/n) + v+ \sigma^2}$$
where $v= \mbox{var}(\hat{p})$ is the sampling variance of $\hat{p}$, $v_j$ denotes the sampling variance of $\hat{p}_j$ and $\sigma^2$ equals the between-area variance (i.e., $\mbox{var}_j(p_j)$) \cite{longford1999multivariate}. By plugging the shrinkage estimator into the WOE calculation, we obtain the shrinkage estimator of the WOE values:
$$\widehat{\mbox{SWOE}_j} = \log\left(\frac{\tilde{p}_j }{1-\tilde{p}_j}\right)$$
for each category $j$.
In the rest of the paper, we denote the WOE transformation based on the shrinkage estimation of the proportions by $\swoe (\cdot)$.

In addition to the global shrinkage method described above, which shrinks proportions towards the overall proportion in the data, we consider shrinking the proportions locally. More specifically, we cluster the WOE values using the weighted 
%Editor: Please note that both ``k`` and ``K`` have been used in the manuscript. Either is acceptable, but please consider using only one consistently throughout.
$k$
-means approach \cite{macqueen1967some, lloyd1982least}, where the weights are taken as inversely proportional to the sampling variability of the WOE values. Note that by the central limit theorem and delta method, it holds that $\sqrt{n} (g(\hat{p}) - g(p)) \xrightarrow{D}N\left(0, \frac{1}{p(1-p)}\right)$, where $g(t) = \log\left(\frac{t}{1-t}\right)$. The asymptotic variance of the WOE estimates is thus $1/(np(1-p))$. Denoting the $\woe$ values with $z_1, \ldots, z_n$, we therefore solve the optimization problem given by
$$\hat{B}_1, \ldots, \hat{B}_K  =\argmin_{B_1,\ldots,B_K}{\sum_{k = 1}^{K}\sum_{i \in B_k}{w_i(z_i - \bar{z}_k)^2}}$$
where $w_i \sim n_{j_i}\hat{p}_{j_i}(1-\hat{p}_{j_i})$ and $j_i$ is the category of the original observation $x_i$. Note that these weights are small for categories with very few observations, which makes it more likely that these categories are put in the same cluster as other categories. Clustering the WOE values induces local shrinkage, since WOE values that are close together tend to end up in the same cluster and receive a WOE value that is a weighted average of the WOE values in the cluster. In addition to achieving less variability in the estimation of the WOE values, we also obtain a natural ``fusing'' of similar categories resulting in a categorical variable with fewer categories. This allows for easier interpretation and visualization of the effect of the categorical variable. In the rest of the paper, we denote the WOE transformation based on clustered estimation of the proportions by $\cwoe (\cdot)$.

To allow for nonlinear effects of the predictor variables on the log-odds, we revisit the approach of \cite{Henckaerts2018} and start from the generalized additive model (GAM) of Eq.~\ref{eq:gam}. After fitting the GAM, the goal is to discretize the fitted spline functions into a limited number of bins. These can then be used as inputs for a classical logistic regression model. As such, we can capture nonlinear effects while greatly improving the interpretability of the model.\par
Our approach differs from others in three main ways. First, we unite different binning types in one framework consisting of ``constrained'' and ``unconstrained'' binning. Both types have the same elegant objective function (with an additional constraint in the former case), which can be optimized exactly and efficiently. Second, we avoid the use of evolutionary trees for constrained binning, as they are typically slow to compute and do not guarantee a global optimum of the objective function. Finally, our framework allows for a natural inclusion of weights in both types of binning, and these are typically chosen to be inversely proportional to the variance of the estimated spline function at the observed value. This strategy avoids creating too many bins in those regions of the spline function which are supported by only a few observations. 

\noindent Depending on the nature of the predictor variable, different types of binning may be desirable. We distinguish two cases:
\begin{enumerate}
\item \textbf{Unconstrained} binning: the value of the original feature does not play a role in the binning process.
\item \textbf{Constrained} binning: the value of the original feature imposes a monotonicity constraint on the binning process.
\end{enumerate}

Let us consider an example. Suppose that $x_j$ is a variable characterizing the age of a person making a transaction. After fitting the model in Eq.~\ref{eq:gam}, we obtain a smooth function $f_j(x_j)$ that linearly influences the log-odds. Suppose that we want to create bins for this transformed variable. If we apply unconstrained binning, the binning of $f_j(X_j)$ would be independent of the value of $X_j$. This means that the resulting bins may combine different age groups. We could have a bin of ages $\{0-20, 80+\}$ and another bin of ages $\{21-79\}$. While this may be fine in some situations, there may also be situations where the binning process is required to be contiguous in $x_j$ to enable a user to interpret or explain the model. This means that the categories cannot ``jump'' over ages. An example of such a binning result is $\{0-50\}$ and $\{50+\}$. We would like to emphasize that the choice of binning is primarily a question of user preferences.\\

\noindent Unconstrained binning is arguably the easiest problem. Given a predictor $\bx = x_1,\ldots,x_n$ where $i = 1,\ldots,n$ ranges over the observations, consider the transformed values $z_i =  f(x_i)$. We want to find $K$ disjoint bins $\hat{B}_1, \ldots, \hat{B}_K$ for the original observations $x_1, \ldots, x_n$ such that within each bin, the corresponding values of $z_i$ are roughly homogeneous. This is a univariate clustering problem for which many approaches have been proposed.
We propose to optimize the weighted $k$-means objective function:

$$\hat{B}_1, \ldots, \hat{B}_K  =\argmin_{B_1,\ldots,B_K}{\sum_{k = 1}^{K}\sum_{i \in B_k}{w_i(z_i - \bar{z}_k)^2}}$$
where $w_i \geq 0$ are weights such that $\sum_{i=1}^{n}{w_i} = n$ and $\bar{z}_k$ denotes the mean of all $z_i$ values with $i \in B_k$ (i.e., the cluster center). We choose the weights to be inversely proportional to the variance of the fitted spline function at point $x_i$. Once we obtain the bins $\hat{B}_1, \ldots, \hat{B}_K$, we can transform the original predictor $\bx = x_1,\ldots,x_n$ to $\bar{z}_{k_1},\ldots, \bar{z}_{k_n}$, where $k_i$ denotes the cluster to which observation $i = 1,\ldots,n$ is assigned. Alternatively, we can include the predictor as a categorical variable with the categories equal to the cluster memberships. We choose not to do this to avoid the creation of many dummy variables.\par

The weighted $k$-means clustering problems can be solved exactly in $\mathcal{O}(n\log(n))$ time using dynamic programming. Finally, note that the $k$-means approach with all weights equal to 1 is equivalent to Fisher's natural breaks algorithm \cite{Fisher1958} used in \cite{Henckaerts2018}. The issue of choosing the number of clusters $K$ is a challenge in cluster analysis, and a multitude of heuristic approaches exist. Among the more popular methods are the gap statistic \cite{Tibshirani2001} and the silhouette coefficient \cite{Rousseeuw1987}. While these can be used in our setting, they do not take our primary goal of building a solid predictive model into account. Therefore, we adopt a hyperparameter tuning approach and determine the value of $K$ by evaluating the quality of the resulting logistic regression model, aligning the clustering process with our overall objective. We address this issue in more detail in Section~\ref{sec:modelbuilding}.\par

We now turn to the problem of constrained binning. Consider again the transformed variable $z_i =  f(x_i)$. In contrast to the unconstrained binning scenario, the value of $x_i$ now influences the clustering of the $z_i$ values. Suppose without loss of generality that the values of $x_i$ are ordered in the relevant order (e.g., the observed ages are listed in ascending order). We are now interested in $K$ bins $B_1, \ldots, B_K$, which each contain disjoint subsets of $x_1,\ldots, x_n$ such that if $x_i, x_j \in B_k$ for certain $i < j \in \{1, \ldots, n\}$, then $x_l\in B_k$ for all $i \leq l \leq j$. Of course, we still want the bins to contain homogeneous values for the corresponding transformed values $z_i$. This problem is equivalent to fitting a step function to the set of bivariate points $(x_i, z_i)$, i.e., we look for a piecewise-constant approximation of $z_i$ within the clusters of $x_i$. This problem has been considered in many areas, including function approximation, time series analysis and cluster analysis. In the same spirit as the weighted $K$-means approach, we propose to optimize the weighted $K$-segments objective function:
$$\hat{B}_1, \ldots, \hat{B}_K  =\argmin_{B_1,\ldots,B_K}{\sum_{k = 1}^{K}\sum_{i \in B_k}{w_i(z_i - \bar{z}_k)^2}}$$
which is the exact same objective as that of the weighted $k$-means problem, with the added constraint that the bins need to be contiguous. The weights $w_i \geq 0$ are again chosen to be inversely proportional to the variance of the fitted function at point $x_i$.

The $k$-segments clustering can be found exactly in $\mathcal{O}(n^2)$ time using dynamic programming, but an approximate $\mathcal{O}(n\log(n))$ algorithm exists \cite{wang2011ckmeans}. Alternatively, one could use (evolutionary) regression trees to bin the $z_i$ values. However, they are typically slower to compute and do not guarantee a global optimum of the objective function.

\subsection{Model building}\label{sec:modelbuilding}

We now discuss how to incorporate the new techniques when building a GLM.
For each continuous effect that is discretized into a step function, there is one tuning parameter in the form of the number of bins used. For categorical data, the global shrinkage estimation of the WOE values does not have additional tuning parameters, but when using clustering to achieve local shrinkage, the number of clusters is a tuning parameter. Ideally, one would optimize a performance criterion of choice over all possible combinations of the tuning parameters, but this evidently becomes computationally cumbersome when there are multiple nonlinear continuous variables and clustered categorical variables.\par
We propose to simplify the problem as follows. A simple 
%Editor: Abbreviations and acronyms are typically defined the first time the term is used within the main text and then used throughout the remainder of the manuscript. Please consider adhering to this convention. The target journal may have a list of abbreviations that are considered common enough that they do not need to be defined.
AIC 
for univariate $k$-means clustering is \cite{ramsey2008uncovering} $\mbox{AIC} = \mbox{WCSS}_k + 2 k$, where $k$ equals the number of clusters and $\mbox{WCSS}_k$ denotes the within-cluster sums of squares (i.e., the value of the $k$-means objective) when clustering into $k$ clusters. One could use this formula to select the number of clusters for each clustering problem, but this would not take the performance of the final model into account. Therefore, we adapt this criterion by introducing a parameter that balances the strength of the fit with the number of clusters:
\begin{equation}\label{eq:tuningobj}
\mbox{WCSS}_k + \lambda k.
\end{equation}
As $\lambda$ increases, we encourage the algorithm to use fewer bins or clusters. When there are only continuous variables that need to be preprocessed, we propose to use two tuning parameters, $\lambda_{c}$ and $\lambda_{uc}$, for the constrained and unconstrained effects, respectively. Note that it is necessary to distinguish between these two effects since the constrained problem will have a naturally higher $\mbox{WCSS}$. To tune the model, we thus use the procedure outlined in Table~\ref{Table 1}.

\begin{table}[!ht]
\centering
\begin{tabular}{|p{2cm}|p{9cm}|}
\hline
\multicolumn{2}{|c|}{Tuning of $\lambda_{c}$ and $\lambda_{uc}$}\\
\hline \hline
Step 1 & Bin the unconstrained nonlinear continuous effects using the number of bins $k$ that yields the minimal value of the objective in Eq.~\ref{eq:tuningobj} with $\lambda = \lambda_{uc}$.\\
Step 2 & Bin the constrained nonlinear continuous effects using the number of bins $k$ that yields the minimal value of the objective in Eq.~\ref{eq:tuningobj} with $\lambda = \lambda_{c}$.\\
Step 3 & Fit a GLM using the binned effects, possibly including other variables.\\
Step 4 & Evaluate the GLM using the AIC.\\
\hline
\end{tabular}
\caption{{\bf Tuning strategy for the binning of the splines.}}
\label{Table 1}
\end{table}

Finally, we choose the combination of tuning parameters yielding the lowest AIC value. This procedure can be used in combination with the shrinkage estimation of the WOE values since the latter procedure does not have a tuning parameter. If the WOE values need to be clustered as well, there is one additional tuning parameter $\lambda_{cat}$. In that case, this parameter is optimized first, as the nature of an effect (linear vs. nonlinear) may change after clustering the WOE values. For each value of $\lambda_{cat}$, we thus execute the procedure outlined in Table~\ref{Table 2}, after which the value of $\lambda_{cat}$ yielding the lowest AIC of the resulting GAM is retained.

\begin{table}[!ht]
\centering
\begin{tabular}{|p{2cm}|p{9cm}|}
\hline
\multicolumn{2}{|c|}{Tuning of $\lambda_{cat}$}\\
\hline \hline
Step 1 & For each categorical variable, find the number of clusters associated with the value of $\lambda = \lambda_{cat}$.\\
Step 2 &Cluster the WOE values of all categorical variables using the appropriate number of clusters found in the previous step.\\
Step 3 & Train the GAM using the splines for the continuous variables and the clustered WOE values for the categorical variables.\\
Step 4 & Evaluate the GAM using the AIC.\\
\hline
\end{tabular}
\caption{{\bf Tuning strategy for the clustering of the WOE values of categorical variables.}}
\label{Table 2}
\end{table}

Once $\lambda_{cat}$ has been determined, we proceed by tuning $\lambda_{c}$ and $\lambda_{uc}$ using the previous procedure in Table~\ref{Table 1}.
Note that instead of using the AIC, the tuning parameters can also be tuned using other performance criteria, such as a measure of prediction accuracy, on a validation set (if available) or through cross validation. This requires more data to be available and more computation time but is likely to better guard against the overfitting of the training data. The parameters $\lambda_{co}$ and $\lambda_{ca}$ yielding the lowest out-of-sample prediction errors are then retained, and the final model is fit using these values.\\

We now analyze the computational complexity of the whole pipeline including the tuning procedure. Suppose the data consists of $n$ observations in $p$ dimensions in addition to a univariate response. Furthermore, assume that the continuous variables can be split up in $p_{uc}$ unconstrained nonlinear effects, $p_c$ constrained nonlinear effects, and $p_l$ linear effects. Also denote the number of categorical variables with $p_{cat}$ so that $p = p_{uc} + p_c + p_l + p_{cat}$. Finally, We assume that the lengths of the grids for the tuning parameters are given by $G_{cat}$, $G_{uc}$ and $G_c$.\\
The time complexity can now be analyzed by splitting up the procedure in 2 steps, the first being the tuning of $\lambda_{cat}$ for the categorical variables, and the second the tuning of the parameters $\lambda_{uc}$ and $\lambda_{c}$ for the continuous nonlinear effects.\\
Step 1 requires, for each value of $\lambda_{cat}$, the preprocessing of $p_{cat}$ variables and the fitting of one GAM on $\bX$. The preprocessing requires $\mathcal{O}(n\log(n))$ time for each categorical variable  due to the exact univariate $k$-means optimization. In total, we thus obtain $\mathcal{O}(G_{cat}\left(C_{GAM}+p_{cat} n\log(n)\right))$, where $C_{GAM}$ denotes the computational cost of fitting a GAM to the data. Note that step 1 is only needed when the WOE values need to be binned. In case shrinkage estimation is used, there is no need for the tuning parameter $\lambda_{cat}$ and the complexity becomes $\mathcal{O}(C_{GAM})$. The complexity of fitting a GAM depends on the fitting algorithm and the number of smoothing parameters, but $\mathcal{O}(np^2)$ is a reasonable assumption given a fixed number of iterations until convergence (see \cite{wood2017generalized,li2020faster} for a discussion).\\
Step 2 requires the separate preprocessing of the constrained and unconstrained effects through spline-binning. This requires $\mathcal{O}(\left(G_cp_c + G_{uc}p_{uc}\right)n\log(n))$ time. Additionally, for each combination of  $\lambda_{uc}$ and $\lambda_{c}$, the fitting of one GLM is required, which leads to an additional $\mathcal{O}(G_c G_{uc}np^2)$ cost.\\
Combining the computational cost of both steps together, we obtain a total of $\mathcal{O}(G_{cat}\left(np^2+p_{cat} n\log(n)\right)) + \mathcal{O}(\left(G_cp_c + G_{uc}p_{uc}\right)n\log(n))+\mathcal{O}(G_c G_{uc}np^2)$. If we assume the sizes of the grid to be constant for increasing $n$ and $p$, and we further assume that at least one of $p_{uc}, p_c, p_l, p_{cat}$ is $\mathcal{O}(p)$ (which is a worst-case scenario), we obtain an overall complexity of $\mathcal{O}( n\log(n)p + np^2)$. While this is a manageable complexity, the constant factor may be quite high if the grids for parameter tuning are fine. That said, the optimization over a grid can be easily parallelized to allow for efficient yet precise parameter tuning.

\section{Empirical results}

\subsection{Data}

\begin{figure}[!ht]
\includegraphics[width = \columnwidth]{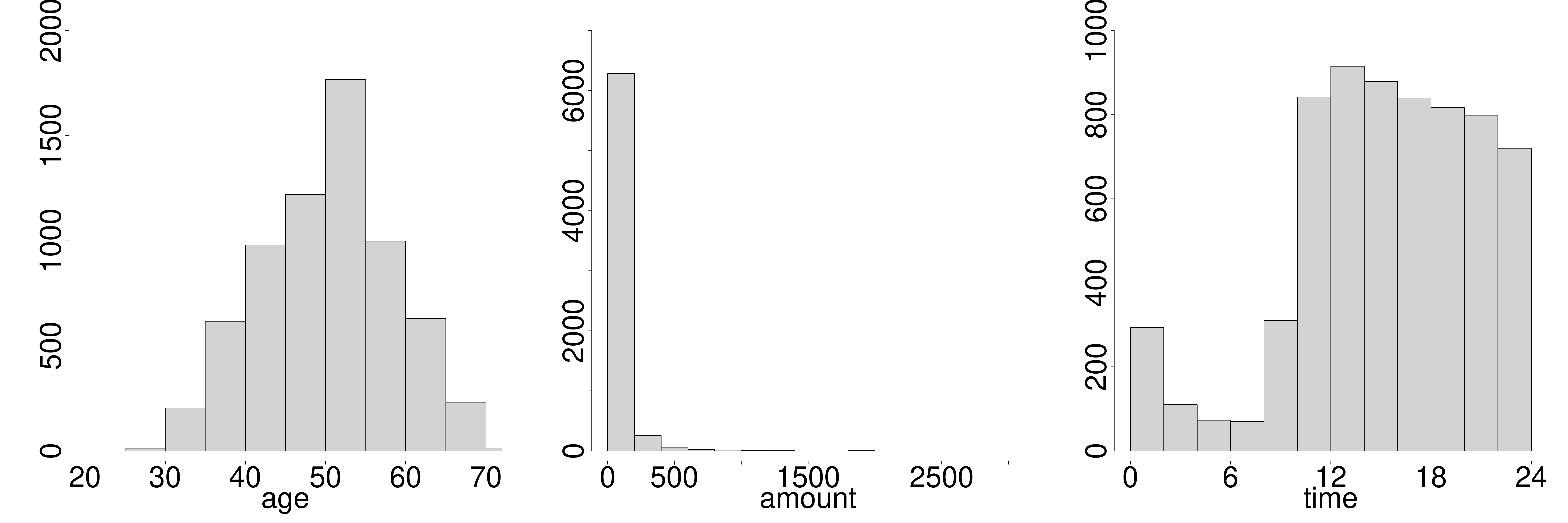}
\caption{{\bf Histograms of the continuous variables in the credit card fraud dataset.}
The \texttt{age} variable (left) is roughly symmetrically distributed, the \texttt{amount} variable is heavily right skewed, and the \texttt{time} variable shows few transactions between 1 and 7 a.m.}
\label{Fig 1}
\end{figure}

We evaluate our proposal on two datasets. The first is a dataset on fraud detection in credit card transactions completed on the east coast of the USA. The dataset consists of training and test sets with 3334 and 3335 points, respectively. For each transaction, 5 variables are recorded: \texttt{amount}, \texttt{age}, risk \texttt{category} (previously assigned by the bank), \texttt{country} and \texttt{time}. The response is a binary variable indicating fraudulent transactions, of which there are 73 in this dataset (i.e., roughly 1 \%). Table~\ref{Table 3} presents an overview of the variables in the dataset, and Fig.~\ref{Fig 1} shows the histograms of the continuous variables.

\begin{table}[!ht]
\begin{center}
 \begin{tabular}{l l} 
 \hline
 Variable name & description \\ [0.5ex] 
 \hline\hline
 \texttt{amount} & transaction amount (USD) \\ 
 \hline
 \texttt{age} & age of the person executing the transaction \\
 \hline
 \texttt{category} & risk category of the transaction (low-medium-high) \\
 \hline
 \texttt{country} & transaction destination (43 countries)\\
 \hline
 \texttt{time} & time of transaction (0-24h) \\ [1ex] 
 \hline
\end{tabular}
\caption{
{\bf Description of the variables in the fraud detection dataset.}}
\label{Table 3}
\end{center}
\end{table}

As a second illustration of our proposal, we use the dataset from the 2009 Pacific-Asia Knowledge Discovery and Data Mining conference (PAKDD) competition. This dataset is about credit risk assessment for private label credit card applications. After removing the constant predictors, we are left with 40000 observations of 20 predictor variables. The response is again binary and indicates whether a credit card application is good or bad, with approximately 20 \% of the applications in the data being bad. The data are publicly available, in, among others, the \texttt{CostCla} Library \cite{package:costCLA}. Of the 20 variables, there are 7 numerical and 13 categorical variables. The names of the categorical variables are listed in Table~\ref{Table 3b} together with the number of categories of each variable. As is clear from this table, there are a number of binary variables but also some variables with multiple categories, including the variable \texttt{PROFESSION\_CODE} with 289 levels.

\begin{table}[!ht]
\begin{center}
\resizebox{0.8\columnwidth}{!}{%
 \begin{tabular}{l c} 
 \hline
 Variable name & Number of categories \\ [0.5ex] 
 \hline\hline
\texttt{ID\_SHOP} & 31\\
\texttt{SEX}& 2\\
\texttt{MARITAL\_STATUS}& 5\\
\texttt{FLAG\_RESIDENCIAL\_PHONE} & 2\\
\texttt{AREA\_CODE\_RESIDENCIAL\_PHONE}& 59\\
\texttt{SHOP\_RANK} & 3\\
\texttt{RESIDENCE\_TYPE} & 4\\
\texttt{FLAG\_MOTHERS\_NAME}& 2\\
\texttt{FLAG\_FATHERS\_NAME}& 2\\
\texttt{FLAG\_RESIDENCE\_TOWN\_eq\_WORKING\_TOWN} & 2\\
\texttt{FLAG\_RESIDENCE\_STATE\_eq\_WORKING\_STATE} &2\\
\texttt{PROFESSION\_CODE} & 289\\
\texttt{FLAG\_RESIDENCIAL\_ADDRESS\_eq\_POSTAL\_ADDRESS} & 2\\
 \hline
\end{tabular}
}
\caption{
{\bf Categorical variables in the credit risk dataset.}}
\label{Table 3b}
\end{center}
\end{table}

\begin{figure}[!ht]
\includegraphics[width = \columnwidth]{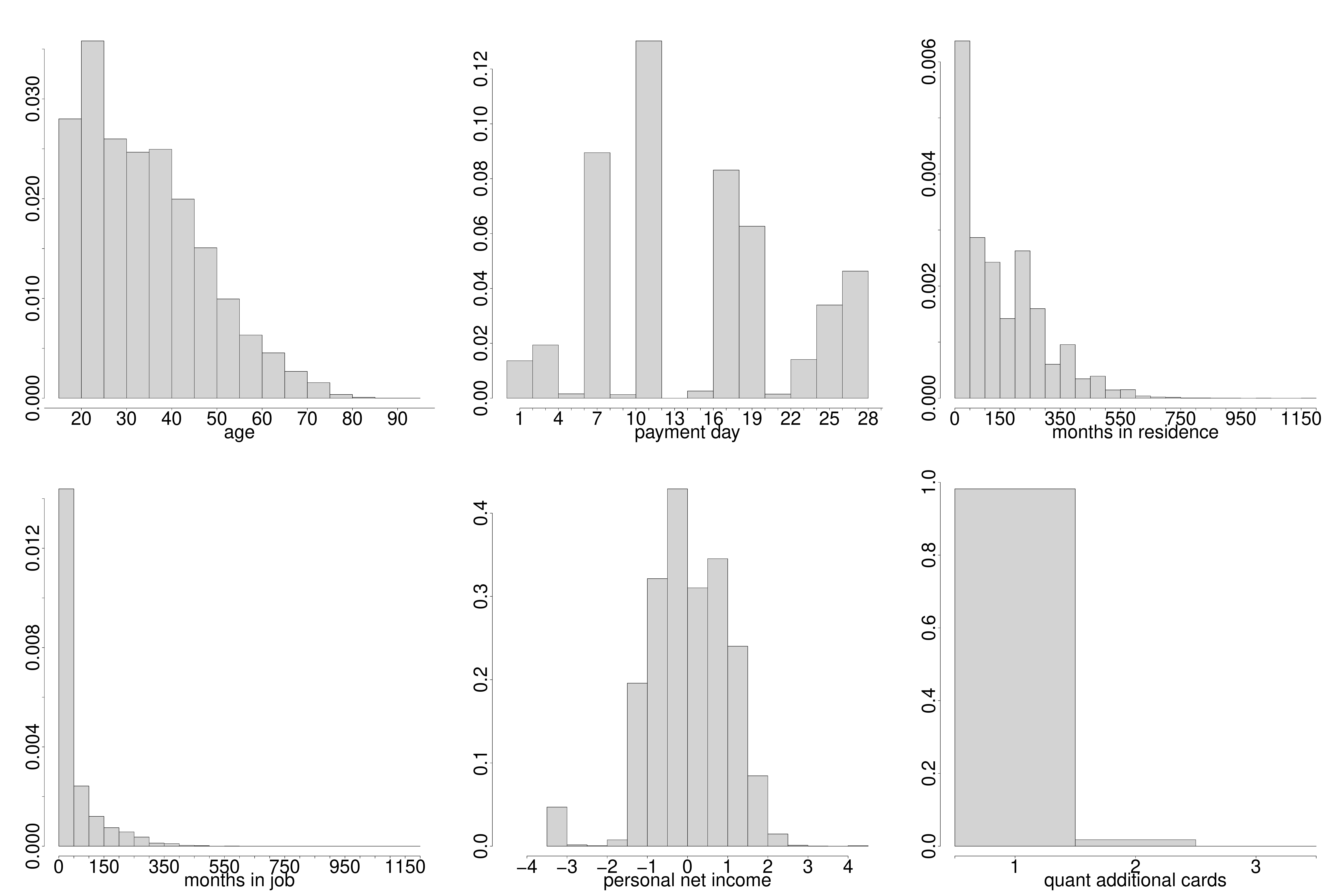}
\caption{{\bf Histograms of the continuous variables in the credit risk dataset.}
The personal net income variable is transformed using a power transformation from the Yeo-Johson family.}
\label{Fig 1b}
\end{figure}

Fig.~\ref{Fig 1b} presents the histograms of the continuous variables in the credit risk dataset, with the exception of the variable \texttt{MATE\_INCOME}, which has over 95\% zeroes and does not allow for an elegant histogram representation. The personal net income variable is transformed towards normality using the Yeo-Johnson power transformation \cite{Yeo2000} fitted by weighted maximum likelihood \cite{raymaekers2021transforming}.

\clearpage
\subsection{Experimental design}
To illustrate the advantages of the proposed method in several ways, we set up three experiments. The first two are conducted on the credit card fraud data and are meant to illustrate the model building process step-by-step while emphasizing the enhanced interpretability and superior results of the resulting model. The third experiment is a complete comparison of the proposed method on the credit risk data using cross validation. For our experiments, we make use of the $\texttt{R}$ packages \texttt{mgcv} \cite{wood2012mgcv}, \texttt{Ckmeans.1d.dp} \cite{wang2011ckmeans}, \texttt{cellWise} \cite{cellWise}, \texttt{hmeasure} \cite{hmeasure}, \texttt{xgboost} \cite{xgboost}, \texttt{caret} \cite{caret} and \texttt{ROCR} \cite{rocr}.

\subsubsection{Experiment 1: the effect of spline binning on the fraud dataset}
\begin{figure}[!h] % Fig 2
\includegraphics[width = \columnwidth]{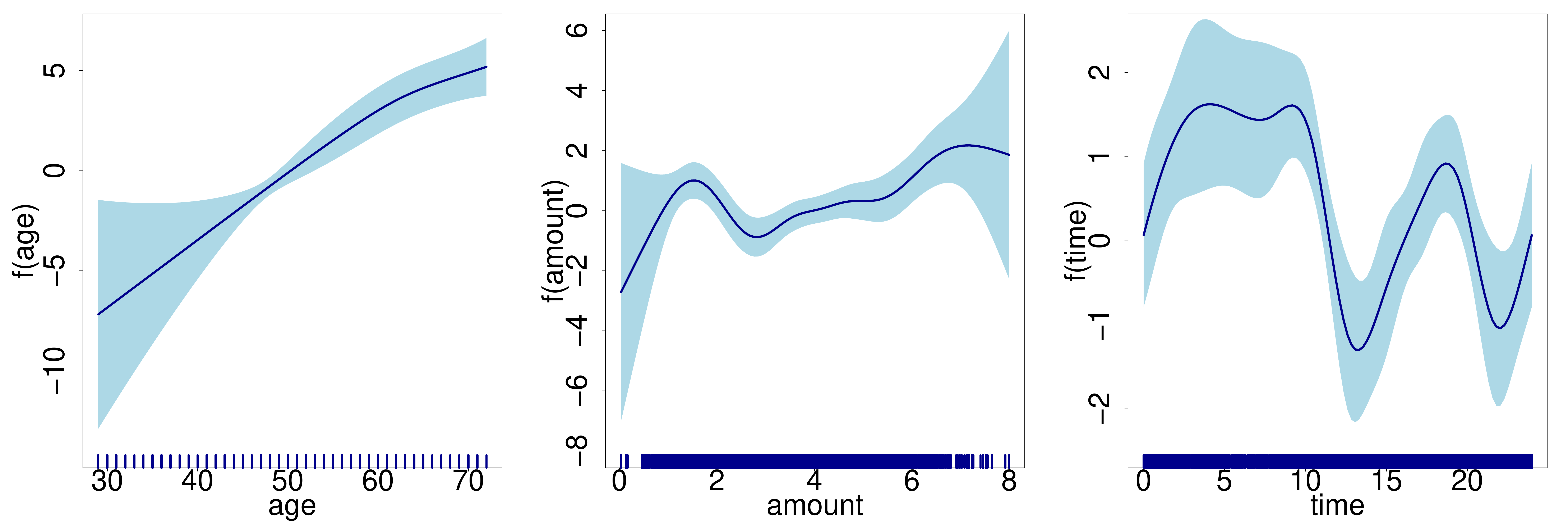}
\caption{{\bf Results of a classical GAM fit to the continuous predictors.}
The fitted splines suggest a quasi-linear effect for the \texttt{age} variable (left) and nonlinear effects for the \texttt{amount} (middle) and \texttt{time} (right) variables on the log-odds.}
\label{Fig 2}
\end{figure}

In the first experiment, we use only the continuous variables to predict fraudulent transactions. To quickly scan for the variables that may have potential nonlinear effects on the response, we fit a GAM on the continuous predictors. Fig.~\ref{Fig 2} shows the results, indicating that the \texttt{amount} and \texttt{time} variables are likely to influence the log-odds of fraud in a nonlinear way. Note that the \texttt{time} variable is a typical example of an inherent nonlinear effect, as discussed in Section~\ref{sec:meth}.

Denoting by $p$ the probability of fraud, we train the following GAM on the training data:
\begin{equation}\label{eq:gamfraud}
 \log\left(\frac{p}{1-p}\right) = \beta_0 + \beta_1 \mbox{\texttt{age}} +  f_1(\mbox{\texttt{amount}}) +f_2(\mbox{\texttt{time}}) 
\end{equation}
where $f_1$ is a thin-plate regression spline \cite{WoodThin} and $f_2$ is a cyclic cubic regression spline \cite{WoodGam}, which captures the periodic nature of the time effect.

In the second step, the continuous effects $f_1(\mbox{\texttt{amount}})$ and $f_2(\mbox{\texttt{time}})$ are discretized (i.e., approximated by step functions) using the strategy described in Section~\ref{sec:modelbuilding} to obtain \texttt{f(amount)} and \texttt{f(time)}. The \texttt{amount} variable is discretized using constrained binning, whereas we use unconstrained binning for the \texttt{time} variable.\\
Finally, a classical logistic regression model is fit to the transformed variables:
\begin{equation*}\label{eq:glmfraud}
 \log\left(\frac{p}{1-p}\right) = \beta_0 + \beta_1 \mbox{\texttt{age}} +  \beta_2 \mathrm{f}_1\texttt{(amount)} + \beta_3 \mathrm{f}_2\texttt{(time)} 
\end{equation*}

The results are evaluated based on different criteria. In addition to the AIC on the training set, we also evaluate the AUC, the weighted Brier score and the H-measure obtained on the test set. The AUC is the well-known area under the receiver operating curve (also equivalent to a linearly transformed Gini coefficient). The classical Brier score is the mean squared error between the predicted probabilities and observed responses, i.e., $\frac{1}{n}\sum_{i=1}^{n}{(\hat{p}_i-y_i)^2}$. This measure is clearly inadequate for imbalanced classification tasks, as it gives equal importance to each individual prediction. We therefore use weights that are inversely proportional to the prior probabilities: $\mbox{wbrier} = \frac{1}{n}\sum_{i=1}^{n}{w_i(\hat{p}_i-y_i)^2}$, where $w_i = \frac{1}{\pi_0} I_{y_i = 0} + \frac{1}{\pi_1} I_{y_i = 1}$. Note that these weights make the predictions of all fraudulent cases together as important as those of all regular transactions. The H-measure is a more recently developed alternative to the AUC that avoids dependence on the classifier and is therefore more reliable. It requires the severity ratio as an input, for which we take the recommended ratio of the class priors ($\pi_1 / \pi_0$); see \cite{hand2009, hand2010} for details.

\subsubsection{Experiment 2: complete approach on the fraud dataset}
In the second experiment, we consider the complete fraud dataset (including the categorical variables) with the goal of evaluating the different treatment combinations of the categorical and continuous variables. For the combination of discretized splines with the shrinkage estimation of the WOE values, we first convert the categorical variables into continuous variables using shrinkage estimators. Then, we proceed as in Experiment 1, with the difference being that the GAM now includes the transformed categorical variables:
\begin{align*}
 \log\left(\frac{p}{1-p}\right) &= \beta_0 + \beta_1 \mbox{\texttt{age}} + \beta_2 \swoe(\mbox{\texttt{category}})\\
&+ \beta_3 \swoe(\mbox{\texttt{country}}) + f_1(\mbox{\texttt{amount}})\\
&+f_2(\mbox{\texttt{time}}) 
\end{align*}
where $f_1$ is a thin-plate regression spline and $f_2$ is a cyclic cubic regression spline, which captures the periodic nature of the time effect.\par

For the combination of the clustered WOE values with the discretized splines, we follow the strategy outlined in Section~\ref{sec:modelbuilding}. We thus first optimize the number of clusters for each of the categorical variables using the approach in Table~\ref{Table 2}. Afterwards, we proceed as in Experiment 1 but now with the GAM:
\begin{align*}
 \log\left(\frac{p}{1-p}\right) &= \beta_0 + \beta_1 \mbox{\texttt{age}} + \beta_2 \cwoe(\mbox{\texttt{category}})\\
&+ \beta_3 \cwoe(\mbox{\texttt{country}})+ f_1(\mbox{\texttt{amount}})\\
&+f_2(\mbox{\texttt{time}}) 
\end{align*}
For the evaluation, we use the same performance measures as in the previous experiment: the AIC, AUC, weighted Brier score and H-measure.

\subsubsection{Experiment 3: complete approach on the credit risk dataset}
In this experiment, we use the same approach as in Experiment 2 in that we compare the combinations of spline binning with the different treatments of categorical variables. We again use the strategy outlined in Section~\ref{sec:modelbuilding}, including the clustering of the categorical variables as in Table~\ref{Table 2} when cWOE is used. All of the continuous variables are fit as a binned spline in the model, with the exception of \texttt{QUANT\_ADDITIONAL\_CARDS}, as it is supported on a very discrete domain. The \texttt{PAYMENT\_DAY} variable, which indicates the day of the month on which the eventual payments will be made, is fitted with a cyclic spline, as it is natural to expect cyclic behavior from this variable. As there is no predefined split for the training and test data, we evaluate our proposal using 10-fold cross validation and evaluate the performance of the method on each fold using the AIC, AUC, weighted Brier score and H-measure.

\subsection{Results}
\subsubsection{Experiment 1}
\begin{figure}[!ht] % Fig 3
\includegraphics[width = \columnwidth]{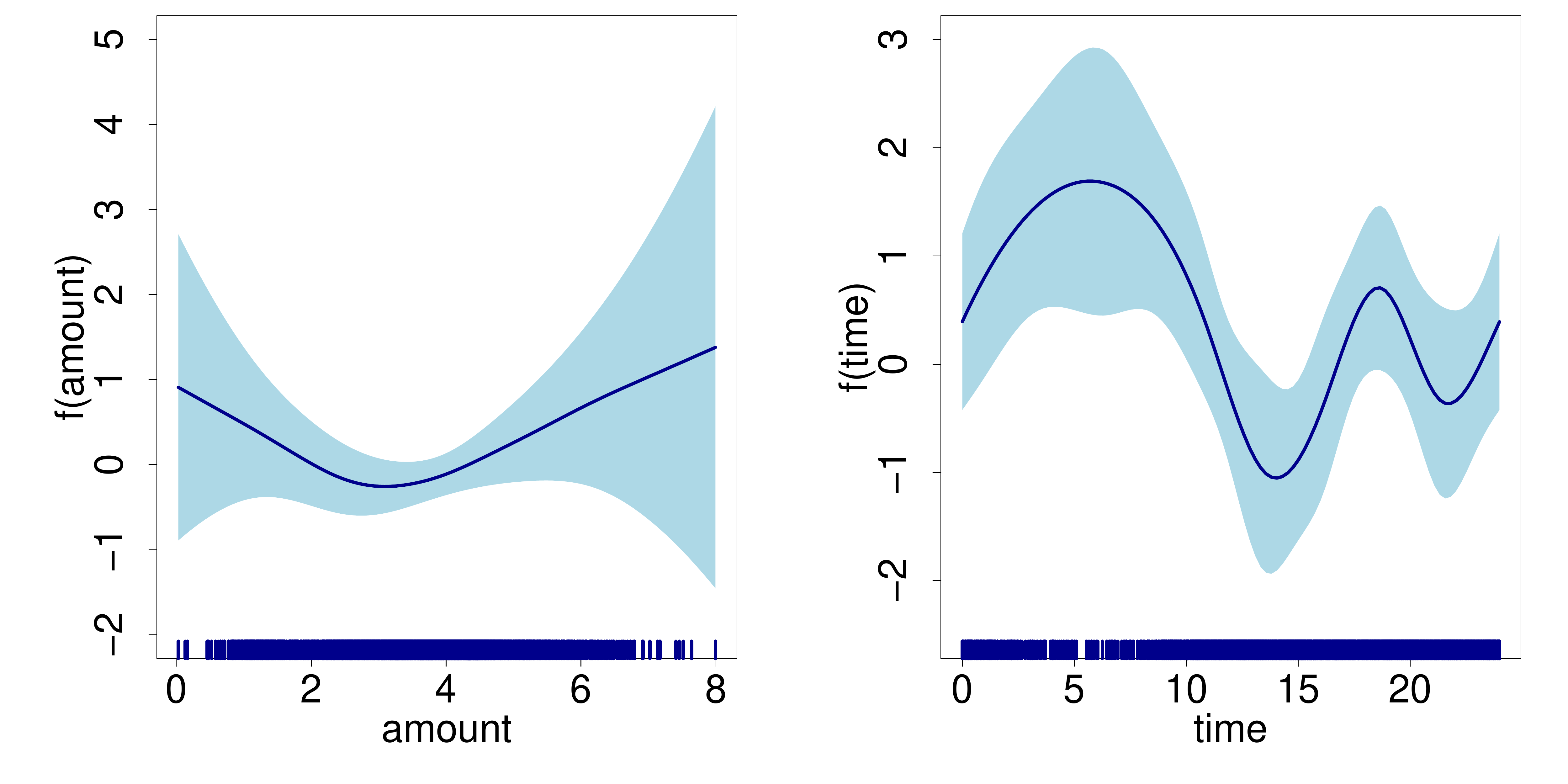}
\caption{{\bf The estimated spline functions of the initial GAM fit for the \texttt{amount} (left) and \texttt{time} (right) variables.}}
\label{Fig 3}
\end{figure}

The initial fit of the GAM of Eq.~\ref{eq:gamfraud} yields the estimates $\hat{\beta}_0 = -19.518$ and $\hat{\beta}_1 = 0.268$, in addition to the spline functions $f_1$ and $f_2$ shown in Fig~\ref{Fig 3}. The fitted amount effect suggests that extreme amounts (both large and small) are more likely to be fraudulent. The time effect suggests that transactions in the morning and late afternoon are more likely to be fraudulent, whereas transactions in the early afternoon and early evening are less likely to be fraudulent. The fitted GAM has an AIC of 284.495. For the out-of-sample measures, we obtain an AUC of 0.919, a weighted Brier score of 0.407 and an H-measure of 0.604. This is a reasonable performance, and we will compare it to the final model and classical GLM later.

\begin{figure}[!ht] % Fig 4
\includegraphics[width = \columnwidth]{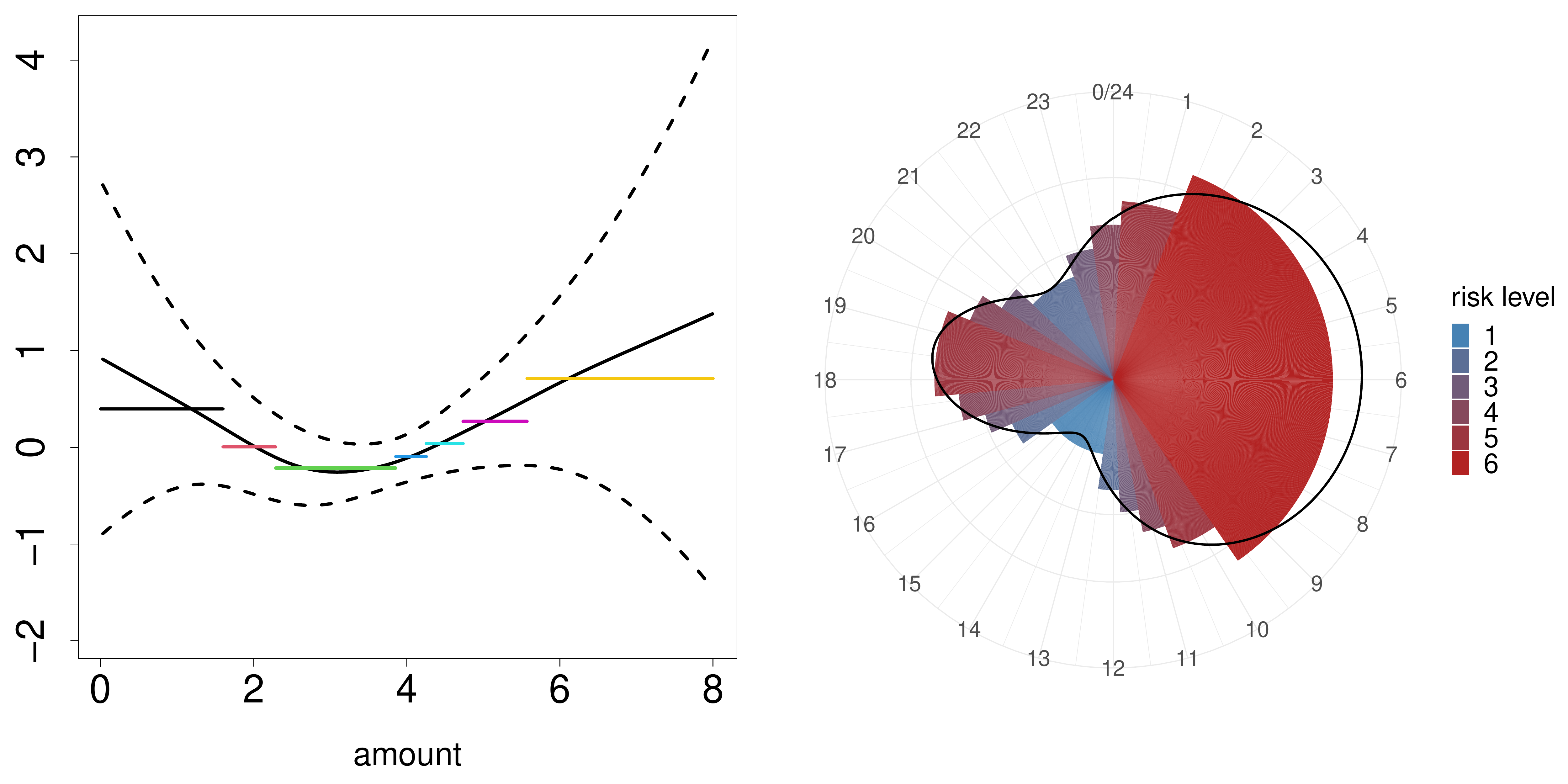}
\caption{{\bf The discretized spline functions of the initial GAM fit for the \texttt{amount} (left) and \texttt{time} (right) variables.}}
\label{Fig 4}
\end{figure}

We now discretize the fitted spline functions. We choose a maximum of $k = 10$ bins and use the selection strategy detailed in Section~\ref{sec:modelbuilding}. This yields 7 bins for the constrained \texttt{amount} binning and 6 bins for the unconstrained binning of the \texttt{time} variable. Fig.~\ref{Fig 4} shows the original and binned effects of both variables. In the left panel, we see the amount variable discretized via a step function with 7 steps. Note that the first and last steps span a rather large interval of transaction amounts. The reason is that there are fewer observations in these regions, and the variance of the estimated spline is much larger. Therefore, due to the weighting strategy with weights inversely proportional to the variances, we obtain larger bins at the extremes of the spline. The right panel shows the time variable, which we wrap around a circle in a clock plot for the purpose of presentation. This plot visually illustrates the time windows in which transactions are more likely to be fraudulent. Note that an effect such as this could never be estimated using classical logistic regression.

We now evaluate the performance of the obtained model using the various performance measures discussed above. The final GLM fit on the discretized splines and the original \texttt{age} variable has an AIC of 286.429. This is slightly above the AIC of the full GAM, but it is clear that the difference is rather small. Furthermore, the tables turn when considering out-of-sample performance. The proposed method yields an AUC of 0.925, a weighted Brier score of 0.396 and an H-measure of 0.624. All of these are in fact better than the corresponding performance measures of the classical GAM fit. This can be explained by the fact that the classical GAM may slightly overfit the training data. By discretizing the resulting spline functions, we gain robustness against this overfitting. Table~\ref{Table 5} shows a comparison of the performances. We additionally add the results of the classical GLM. We see that the GLM with spline binning (SB) outperforms the classical GLM on all levels. The most significant difference is found in the H-measure, with an increase of almost 15 \%. As a reference, we add the performance of XGBoost (XGB) \cite{10.1145/2939672.2939785} to the table, which does not provide a significant improvement over the GLM-based approaches on these data.

\begin{table}[!ht]
\begin{center}
 \begin{tabular}{l rrrr} 
 \hline
 Method & AIC & AUC & wbrier & H-measure\\ [0.5ex] 
 \hline\hline
 \texttt{classical GLM} & 293.656  & 0.896  &0.438    & 0.549  \\ 
 \hline
 \texttt{classical GAM} & \textbf{284.495} & 0.919 &  0.407 & 0.604 \\
 \hline
 \texttt{SB GLM} &  286.429 & \textbf{0.925} &  \textbf{0.396}   &  \textbf{0.624} \\
 \hline \hline
 \texttt{XGBoost} &  NA & 0.891 &    0.363 &  0.567  \\
 \hline
\end{tabular}
\end{center}
\caption{{\bf Comparison of the different models trained on the continuous predictors of the fraud detection dataset.}The GLM with spline binning (SB) outperforms the other methods in the out-of-sample evaluation, whereas the classical GAM has a slightly lower AIC.
}
\label{Table 5}
\end{table}

\subsubsection{Experiment 2}
In the second experiment, we compare the different combinations of our proposed preprocessing techniques. The results of this comparison are presented in Table~\ref{Table 6}. Several interesting conclusions can be made from these results. First, we see that the classical GLM is vastly outperformed by any of the other methods. This is mainly due to the inclusion of 42 dummy variables for the categorical variable \texttt{country}. Second, we can see that the shrinkage estimation of the WOE values outperforms the classical WOE, regardless of whether the continuous effects are estimated using discretized splines. The clustered WOE values do not significantly outperform the classical WOE values, and their main benefit thus lies in the fact that the final model is more interpretable, since it enforces a natural reduction in the number of categories within the categorical variables. Finally, we see that the discretized spline approach always improves upon the model obtained using the original continuous variables. The XGB classifier now outperforms the classical GLM but has an inferior performance to that of the GLM approach after preprocessing with WOE.

\begin{table}[!ht]
\begin{center}
 \begin{tabular}{l l l l| llll} 
 \hline
 WOE & sWOE & cWOE & SB & AIC & AUC & wbrier & H \\ [0.5ex]
 \hline\hline 
\emptycb & \emptycb & \emptycb & \emptycb & 285 & 0.831 & 0.366 & 0.520\\
 \hline
\checkedcb & \emptycb & \emptycb & \emptycb & 227 & 0.925 & 0.352 & 0.596 \\
 \hline
\emptycb & \checkedcb & \emptycb & \emptycb & 226  & 0.928 & 0.354 &  0.615 \\
 \hline
\emptycb & \emptycb & \checkedcb & \emptycb & 225  & 0.924 & 0.357 &  0.589 \\
 \hline
\checkedcb & \emptycb & \emptycb & \checkedcb & 217 &  0.941  & \textbf{0.335} & 0.638 \\
 \hline
\emptycb & \checkedcb & \emptycb & \checkedcb & \textbf{216}  & \textbf{0.943} & \textbf{0.336} &  \textbf{0.652} \\
 \hline
\emptycb & \emptycb & \checkedcb & \checkedcb &  219  & 0.936 &  \textbf{0.336}&  0.627 \\
 \hline
\emptycb & \emptycb & \emptycb & XGB & NA  & 0.905 &  0.347 &  0.637\\
 \hline
%\checkedcb & \emptycb & \emptycb & xgb & NA  & 0.918 &  0.364 &  0.611\\
 %\hline
\end{tabular}
\end{center}
\caption{{\bf Evaluation of the combined strategies on the credit card fraud dataset.} The shrinkage estimation of the WOE values in combination with spline binning outperforms the other models. The clustered WOE values in combination with spline binning is the second best-performing model.}
\label{Table 6}
\end{table}

\begin{figure}[!ht] % Fig 5
\includegraphics[width = \columnwidth]{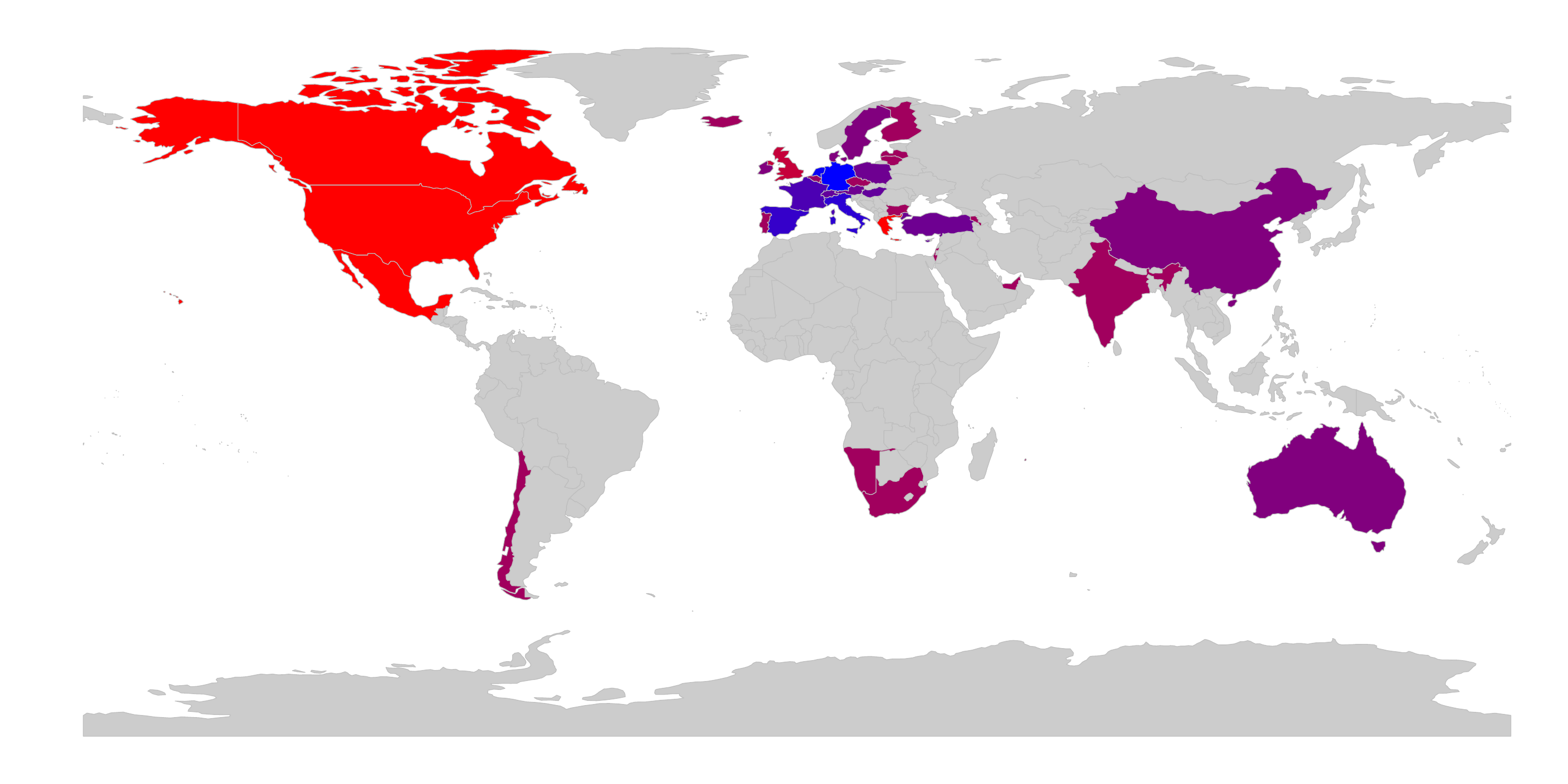}
\caption{{\bf The \texttt{country} variable reduced to 12 categories instead of the original 42.}}
\label{Fig 5}
%\label{fig:finalfit_countries}
\end{figure}

For illustrative purposes, we further analyze the model obtained using clustered WOE values and spline binning.
The clustering of the categorical variables yields an optimal tuning parameter of $\lambda_{\mbox{cat}} = e^{-7}$. This parameter enforces a clustering of the \texttt{country} variable into 12 bins (down from 42 categories), whereas the \texttt{category} variable is left untouched with its original 3 categories. Fig.~\ref{Fig 5} shows the binned \texttt{country} variable with 12 different levels. It turns out that transactions going to Europe are generally connected to lower probabilities of fraud, with the exception being receivers in Greece (and the UK to a lesser extent). The highest risk is associated with national transactions and those to Canada and Mexico. International transactions to Australia, China, South Africa and Chile have neutral risk levels.\\

\begin{figure}[!ht]% Fig 6
\includegraphics[width = \columnwidth]{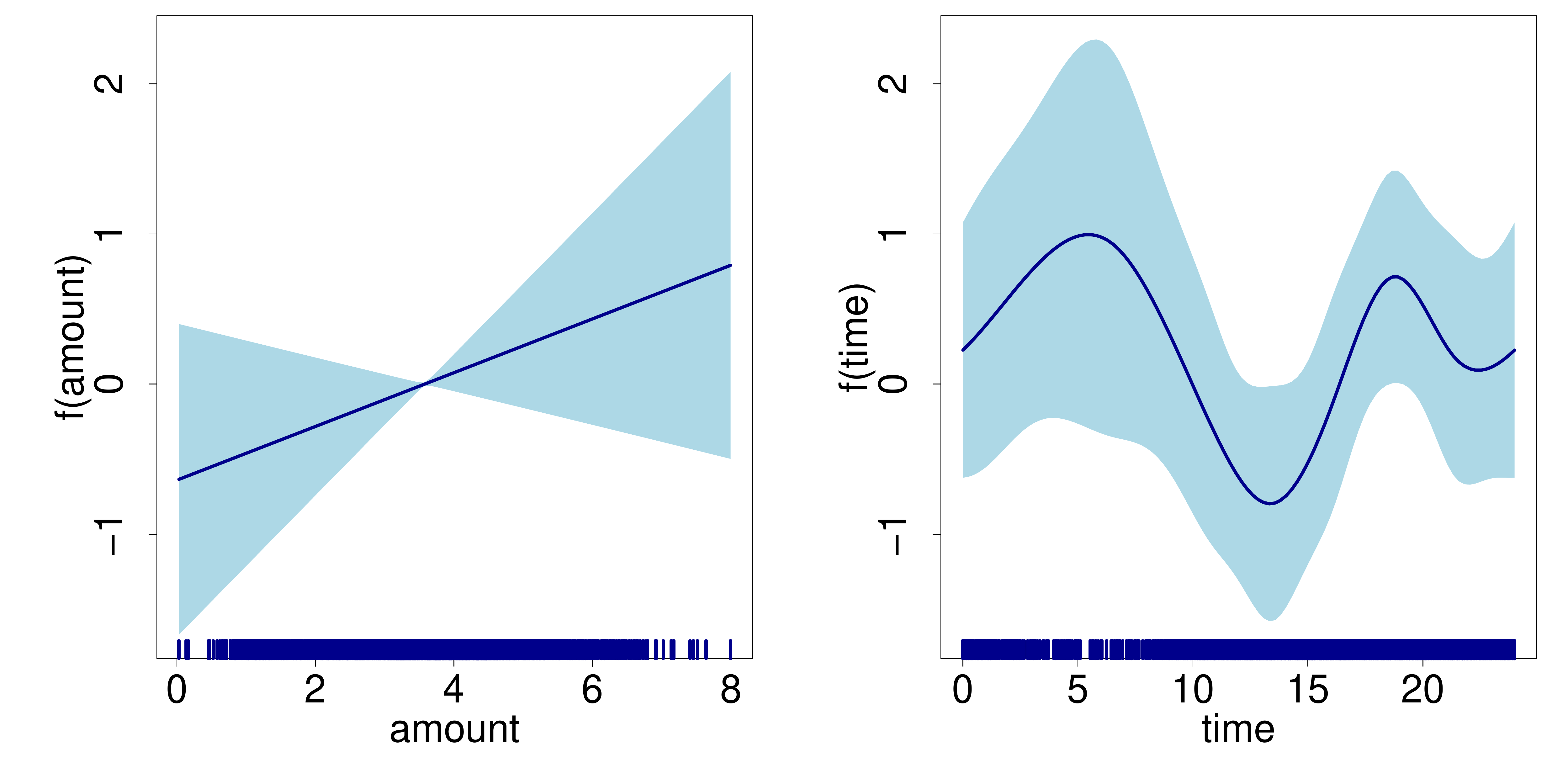}
\centering
\caption{{\bf The estimated spline functions of the initial GAM fit when all variables are included in the model.}
The \texttt{amount} variable (left) no longer displays a nonlinear effect on the response variable, as was the case for the model with only continuous variables.}
\label{Fig 6}
\end{figure}

The GAM fit with the optimal value of $\lambda_{\mbox{cat}}$ no longer displays a nonlinear effect for the \texttt{amount} variable, as was the case in Experiment 1. This means that the inclusion of the categorical variables resolves the nonlinearity issue for this variable, and we can treat it as a linear effect. The \texttt{time} variable, however, still displays a nonlinear relationship with the response, as shown in Fig.~\ref{Fig 6}.

Discretizing the continuous effect of the \texttt{time} variable yields 3 bins. The result of this binning step is shown in Fig.~\ref{Fig 7}. It is clear that the transactions made in the morning or early evening are more likely to be fraudulent than the transactions around noon or late in the evening. The coefficients of the final model are presented in Table~\ref{Table 7}, which suggests that all predictors have significant contributions to the final model, with the exception of the \texttt{amount} variable.

  \begin{minipage}{0.95\textwidth}
  \begin{minipage}[t]{0.49\textwidth}
%\begin{figure}[!ht]%Fig 7
\centering
\raisebox{-1.9cm}{\includegraphics[width = \linewidth]{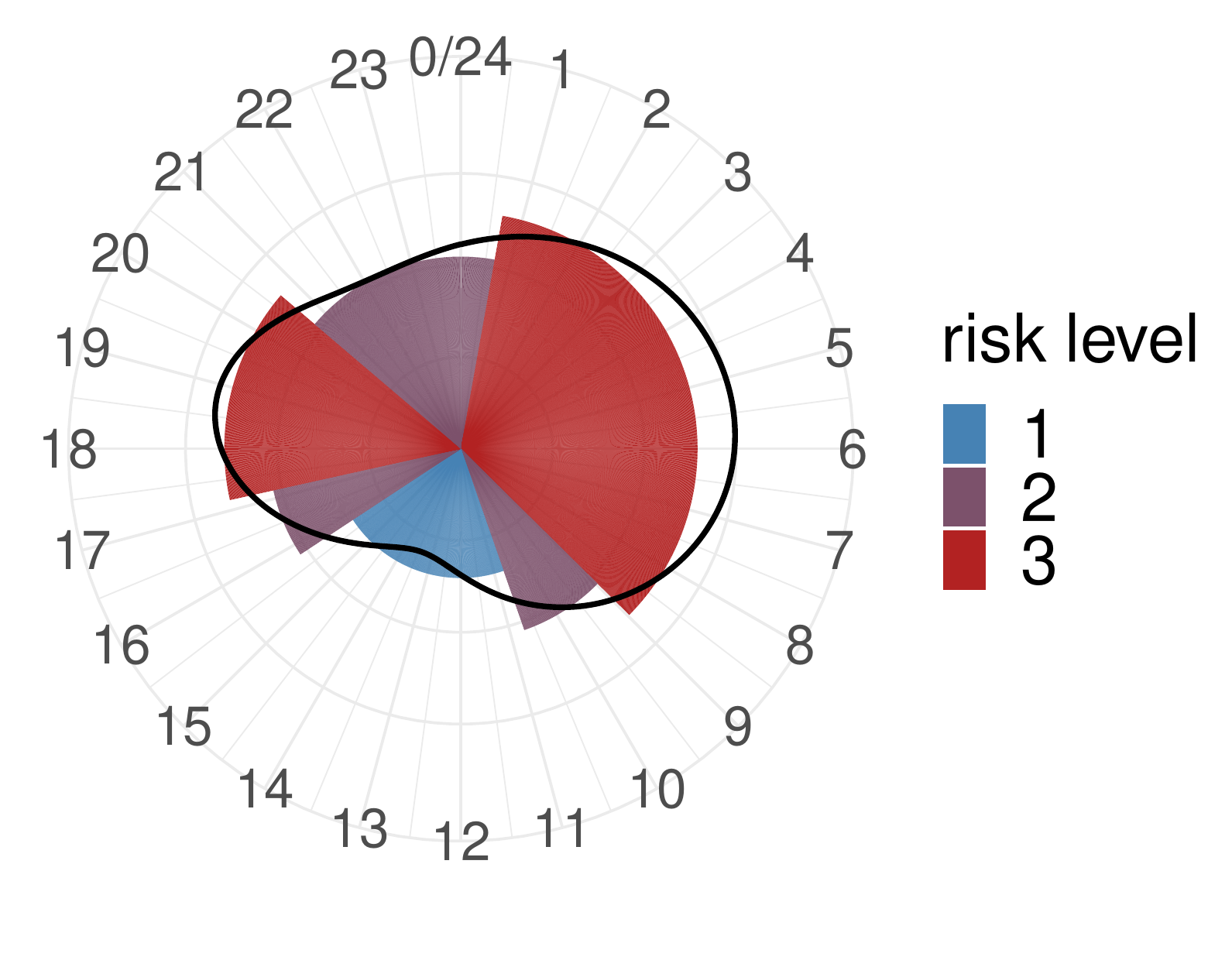}}
\label{Fig 7}
%\end{figure}
\captionof{figure}{{\bf The effect of the binning time on the final model.}}
  \end{minipage}
  \hfill
	\begin{minipage}[t]{0.49\textwidth}
%\begin{table}[ht]
\centering
\resizebox{\linewidth}{!}{
\begin{tabular}{lrr}
  \hline
 & Estimate & P-value \\ 
  \hline
(Intercept) & -12.55 & 0.00 \\ 
    \texttt{amount} & 0.19 & 0.19 \\    \texttt{age} & 0.27 & 0.00 \\   $\cwoe(\texttt{category})$ & 0.64 & 0.01 \\ 
   $\cwoe(\texttt{country})$  & 0.90 & 0.00 \\ 
  \texttt{f(time)}  & 1.88 & 0.00 \\ 
   \hline
\end{tabular}
}
\captionof{table}{{\bf Coefficients of the final model.}}
\label{Table 7}
%\end{table}
 \end{minipage}
  \end{minipage}

\subsubsection{Experiment 3}
The results for the final experiment are summarized in Table~\ref{Table 8}. It is clear that the absolute differences are not as pronounced as those in the previous example. This is not very surprising, as a significant number of predictor variables that carry a lot of signal are either binary or enter the model linearly, and in both cases, the effect of the proposed approach is limited. Nevertheless, all the differences are statistically significant, as verified by the Wilcoxon rank test \cite{wilcoxon1992individual}, which yields p-values between 0.002 and 0.036 for testing the performance of sWOE + SB against the alternatives in terms of the AUC, wbrier and H-measure. These differences can produce significant cost savings in practical business settings. As in the previous example, the XGBoost classifier does not seem to improve upon a GLM-based approach for these data.

\begin{table}[!ht]
\begin{center}
 \begin{tabular}{l l l l| llll} 
 \hline
 WOE & sWOE & cWOE & SB & AIC & AUC & wbrier & H \\ [0.5ex]
 %\hline\hline
%\emptycb & \emptycb & \emptycb & \emptycb & 285 & 0.831 & 0.366 & 0.520\\
 \hline
\checkedcb & \emptycb & \emptycb & \emptycb & 33304.29 &0.6693& 0.3105&    0.1043 \\
 \hline
\emptycb & \checkedcb & \emptycb & \emptycb & 33413.02 & 0.6701 &0.3104 &   0.1056\\
 \hline
\emptycb & \emptycb & \checkedcb & \emptycb & 33305.30& 0.6692& 0.3105 &   0.1045 \\
 \hline
\checkedcb & \emptycb & \emptycb & \checkedcb & \textbf{33184.43}& 0.6732 &0.3087 &   0.1093\\
 \hline
\emptycb & \checkedcb & \emptycb & \checkedcb & 33291.17 &\textbf{0.6746} & \textbf{0.3083} &   \textbf{0.1112}\\
 \hline
\emptycb & \emptycb & \checkedcb & \checkedcb &  33185.57 &0.6733& 0.3086&    0.1096\\
 \hline
\emptycb & \emptycb & \emptycb & XGB & NA & 0.6546 &0.3168 &   0.0886\\
 \hline
\end{tabular}
\end{center}
\caption{{\bf Evaluation of the combined strategies on the credit risk dataset.} The shrinkage estimation of the WOE values in combination with spline binning outperforms the other models.}
\label{Table 8}
\end{table}

\subsection{Discussion}
The results of the experiments above lead us to several conclusions. First, in regard to the estimation of WOE values, estimating the proportions using the shrinkage estimator seems to improve the out-of-sample performance of the resulting model. Second, clustering the WOE values does not generally yield a substantial improvement over the regular WOE values but has the advantage of fusing the categorical variables into a variable with fewer categories, thereby improving the interpretability of the model. Finally, the use of binned splines on the continuous variables significantly improves the out-of-sample performance of the model. Additionally, one could argue that this also leads to improved interpretability, as the continuous variables are reduced to a select number of discrete values. Note that the advantage of using binned splines may not be significant if there are no important nonlinear effects in the set of predictor variables.\par

\section{Conclusion}

We propose and study two advanced techniques for preprocessing data before applying regression. The first method considers the treatment of WOE values, which we propose to estimate using shrinkage estimators for the proportions. Alternatively, the original WOE values can be clustered for improved interpretability. Second, we study the discretization of continuous variables through the binning of spline functions. This allows for capturing nonlinear effects in predictor variables and yields highly interpretable predictors that take only a small number of discrete values.\par
Through three different experiments on a fraud detection dataset, we illustrate the advantages of using these advanced preprocessing techniques. In particular, the out-of-sample performance of the model is improved using the binned spline treatment on the continuous variables. Additionally, the WOE values obtained based on shrinkage estimation of the proportions also increase the out-of-sample performance of the resulting model. The clustering of WOE values shows improved interpretability but no clear improvement in predictive performance.\par
When it comes to the limitations of the proposed method, three points need mentioning.
The first is that it should be possible to adequately model the conditional expectation of the response given the predictors should be appropriately modeled through a generalized additive model. Since this is the starting point of the modeling pipeline, it is a rather obvious yet important limitation. The second is that the computational cost gets quite high when there are many nonlinear continuous effects. As the number of such effects gets higher, GAMs become less and less suitable for modeling. The final limitation is that of risk of overfitting. Whenever GAMs are used, there is the risk of overfitting to the training data, and smoothing parameter selection should be carefully executed. However, there exist reliable automatic routines for this.

Further research could address the combination of the two strategies for categorical variables by using the classical WOE values as inputs for a GAM. This combined method would be able to capture the nonlinear effects of the WOE values on the response. However, due to the nature of WOE in logistic regression (which implies a linear WOE effect on the response), it is not clear that this would yield an improvement over the current method. Another line of research could investigate a more precise approximation of the spline functions in the GAM. For example, one could use a piecewise linear approximation instead of a step function, which would still be easy to interpret but more flexible to work with. Finally, the shrinkage estimation of the proportions could be combined with clustering, i.e., one could first compute WOE values based on shrinkage estimation and then cluster the resulting values in a number of bins.

 \section*{Software availability}
An implementation of the proposed pipeline as well as a script reproducing the results in the paper can be found in the GitHub repository https://github.com/JakobRaymaekers/WOE2.0.

 \section*{Acknowledgments}
	The authors gratefully acknowledge the financial support from the BASF Research Chair on Robust Predictive Analytics, the BNP Paribas Fortis Research Chair in Fraud Analytics at KU Leuven and the Internal Funds KU Leuven under grant C16/15/068. The funders had no role in the study design, data collection and analysis process, the decision to publish, or the preparation of the manuscript.

\bibliography{woe_arxiv}

\end{document}